# Gibbs Sampling in Factorized Continuous-Time Markov Processes

**Tal El-Hay** **Nir Friedman**
School of Computer Science
The Hebrew University
{tale,nir}@cs.huji.ac.il

**Raz Kupferman**
Institute of Mathematics
The Hebrew University
raz@math.huji.ac.il

## Abstract

A central task in many applications is reasoning about processes that change over continuous time. *Continuous-Time Bayesian Networks* is a general compact representation language for multi-component continuous-time processes. However, exact inference in such processes is exponential in the number of components, and thus infeasible for most models of interest. Here we develop a novel Gibbs sampling procedure for multi-component processes. This procedure iteratively samples a trajectory for one of the components given the remaining ones. We show how to perform *exact* sampling that adapts to the natural time scale of the sampled process. Moreover, we show that this sampling procedure naturally exploits the structure of the network to reduce the computational cost of each step. This procedure is the first that can provide asymptotically unbiased approximation in such processes.

## 1 Introduction

In many applications, we reason about processes that evolve over time. Such processes can involve short time scales (e.g., the dynamics of molecules) or very long ones (e.g., evolution). In both examples, there is no obvious discrete "time unit" by which the process evolves. Rather, it is more natural to view the process as changing in a continuous time: the system is in some state for a certain duration, and then transitions to another state. The language of *continuous-time Markov processes* (CTMPs) provides an elegant mathematical framework to reason about the probability of trajectories of such systems (Gardiner, 2004). We consider Markov processes that are homogeneous in time and have a finite state space. Such systems are fully determined by the state space $S$, the distribution of the process at the initial time, and a description of the dynamics of the process. These dynamics are specified by a *rate matrix* $\mathbb{Q}$, whose off-diagonal entries $q_{\boldsymbol{a},\boldsymbol{b}}$ are exponential rate intensities for transitioning from state $\boldsymbol{a}$ to $\boldsymbol{b}$. Intuitively, we can think of the entry $q_{\boldsymbol{a},\boldsymbol{b}}$ as the rate parameter of an exponential distribution whose value is the duration of time spent in state $\boldsymbol{a}$ before transitioning to $\boldsymbol{b}$.

In many applications, the state space is of the form of a product space $S = S_1 \times S_1 \times \cdots \times S_M$, where $M$ is the number of *components* (such processes are called multi-component). Even if each of the $S_i$ is of low dimension, the dimension of the state space is exponential in the number of components, which poses representational and computational difficulties. Recently, Nodelman et al. (2002) introduced the representation language of *continuous-time Bayesian networks* (CTBNs), which provides a factorized, component-based representation of CTMPs: each component is characterized by a conditional CTMP dynamics, which describes its local evolution as a function of the current state of its parents in the network. This representation is natural for describing systems with a sparse structure of local influences between components.

For most applications of such CTMP models, we need to perform inference to evaluate the posterior probability of various queries given evidence. Exact inference requires exponentiation of the rate matrix $\mathbb{Q}$. As the rate matrix is exponential in the number of components, exact computations are infeasible for more than a few components. Thus, applications of factored CTMPs require the use of approximate inference.

In two recent works Nodelman et al. (2005) and Saria et al. (2007) describe approximate inference procedures based on Expectation Propagation, a variational approximation method (Minka, 2001; Heskes and Zoeter, 2002). These approximation procedures perform local propagation of messages between components (or sub-trajectories of components) until convergence. Such procedures can be quite efficient, however they can also introduce a systematic error in the approximation (Fan and Shelton, 2008).

More recently, Fan and Shelton (2008) introduced a procedure that employs importance sampling and particle filtering to sample trajectories from the network. Such a stochastic sampling procedure has anytime properties as collecting more samples leads to more accurate approximation. However, since this is an importance sampler, it has limited capabilities to propagate evidence "back" to influence the sampling of earlier time steps. As a result, when the evidence is mostly at the end of the relevant time inter-val, and is of low probability, the procedure requires many samples. A related importance sampler was proposed by Ng et al. (2005) for monitoring a continuous time process.

In this paper we introduce a new stochastic sampling procedure for factored CTMPs. The goal is to sample random system trajectories from the posterior distribution. Once we have multiple independent samples from this distribution we can approximate the answer to queries about the posterior using the empirical distribution of the samples. The challenge is to sample from the posterior. While generative sampling of a CTMP is straightforward, sampling given evidence is far from trivial, as evidence modifies the posterior probability of earlier time points.

Markov Chain Monte Carlo (MCMC) procedures circumvent this problem by sampling a stochastic sequence of system states (trajectories in our models) that will eventually be governed by the desired posterior distribution. Here we develop a Gibbs sampling procedure for factored CTMPs. This procedure is initialized by setting an arbitrary trajectory which is consistent with the evidence. It then alternates between randomly picking a component $X_i$ and sampling a trajectory from the distribution of $X_i$ conditioned on the trajectories of the other components and the evidence. This procedure is reminiscent of *block Gibbs sampling* (Gilks et al., 1996) as we sample an entire trajectory rather than a single random variable in each iteration. However, in our approach we need to sample a continuous trajectory.

The crux of our approach is in the way we sample a trajectory for a single component from a process that is conditioned on trajectories of the other components. While such a process is Markovian, it is not homogeneous as its dynamics depends on trajectories of its Markov Blanket as well as on past and present evidence. We show that we can perform exact sampling by utilizing this Markovian property, and that the cost of this procedure is determined by the complexity of the current trajectories and the sampled one, and not by a pre-defined resolution parameter. This implies that the computational time adapts to the complexity of the sampled object.

## 2 Continuous-Time Bayesian Networks

In this section we briefly review the CTBN model (Nodelman et al., 2002). Consider an $M$-component Markov process

$$\boldsymbol{X}^{(t)} = (X_1^{(t)}, X_2^{(t)}, \ldots X_M^{(t)})$$

with state space $S = S_1 \times S_2 \times \cdots \times S_M$.

A notational convention: vectors are denoted by boldface symbols, e.g., $\boldsymbol{X}, \boldsymbol{a}$, and matrices are denoted by blackboard style characters, e.g., $\mathbb{Q}$. The states in $S$ are denoted by vectors of indexes, $\boldsymbol{a} = (a_1, \ldots, a_M)$. The indexes $1 \le i, j \le M$ are used to enumerate the components. We use the notation $\boldsymbol{X}^{(t)}$ and $X_i^{(t)}$ to denote a random variable at time $t$. We will use $\boldsymbol{X}^{[s,t]}$, $\boldsymbol{X}^{(s,t]}$, $\boldsymbol{X}^{[s,t)}$, to denote the state of $\boldsymbol{X}$ in the closed and semi-open intervals from $s$ to $t$.

The dynamics of a time-homogeneous continuous-time Markov process are fully determined by the *Markov transition function*,

$$p_{\boldsymbol{a},\boldsymbol{b}}(t) = \Pr(\boldsymbol{X}^{(t+s)} = \boldsymbol{b} | \boldsymbol{X}^{(s)} = \boldsymbol{a}),$$

where time-homogeneity implies that the right-hand side does not depend on $s$. Provided that the transition function satisfies certain analytical properties (continuity, and regularity; see Chung (1960)) the dynamics are fully captured by a constant matrix $\mathbb{Q}$—the *rate*, or *intensity matrix*—whose entries $q_{\boldsymbol{a},\boldsymbol{b}}$ are defined by

$$q_{\boldsymbol{a},\boldsymbol{b}} = \lim_{h \downarrow 0} \frac{p_{\boldsymbol{a},\boldsymbol{b}}(h) - \delta_{\boldsymbol{a},\boldsymbol{b}}}{h},$$

where $\delta_{\boldsymbol{a},\boldsymbol{b}}$ is a multivariate Kronecker delta.

A Markov process can also be viewed as a generative process: The process starts in some state $\boldsymbol{a}$. After spending a finite amount of time at $\boldsymbol{a}$, it transitions, at a random time, to a random state $\boldsymbol{b} \neq \boldsymbol{a}$. The transition times to the various states are exponentially distributed, with rate parameters $q_{\boldsymbol{a},\boldsymbol{b}}$. The diagonal elements of $\mathbb{Q}$ are set such that each row sums up to zero.

The time-dependent probability distribution, $\boldsymbol{p}(t)$, whose entries are defined by

$$p_{\boldsymbol{a}}(t) = \Pr(\boldsymbol{X}^{(t)} = \boldsymbol{a}), \qquad \boldsymbol{a} \in S,$$

satisfies the so-called *forward*, or *master*, *equation*,

$$\frac{d\boldsymbol{p}}{dt} = \mathbb{Q}^T \boldsymbol{p}. \tag{1}$$

Thus, using the $\mathbb{Q}$ matrix, we can write the Markov transition function as

$$p_{\boldsymbol{a},\boldsymbol{b}}(t) = [\exp(t\mathbb{Q})]_{\boldsymbol{a},\boldsymbol{b}},$$

that is, as the $\boldsymbol{a}, \boldsymbol{b}$ entry in the matrix resulting from exponentiating $\mathbb{Q}$ (using matrix exponentiation).

It is important to note that the master Eq. (1) encompasses all the statistical properties of the Markov process. There is a one-to-one correspondence between the description of a Markov process by means of a master equation, and by means of a "pathwise" characterization (up to stochastic equivalence of the latter; see Gikhman and Skorokhod (1975)).

*Continuous-time Bayesian Networks* provide a compact representation of multi-component Markov processes by incorporating two assumptions: (1) every transition involves a single component; (2) each component undergoes transitions at a rate which depends only on the state of a subsystem of components.

Formally, the structure of a CTBN is defined by assigning to each component $i$ a set of indices $\mathrm{Par}(i) \subseteq$

$\{1,\ldots,M\} \setminus \{i\}$. With each component $i$, we associate a *conditional rate matrix* $\mathbb{Q}^{i|\,\mathrm{Par}(i)}$ with entries $q^{i|\,\mathrm{Par}(i)}_{a_i,b_i|\boldsymbol{u}_i}$ where $a_i$ and $b_i$ are states of $X_i$ and $\boldsymbol{u}_i$ is a state of $\mathrm{Par}(i)$. This matrix defines the rate of $X_i$ as a function of the state of its parents. Thus, when the parents of $X_i$ change state, the rates governing its transition can change.

The formal semantics of CTBNs is in terms of a joint rate matrix for the whole process. This rate matrix is defined by combining the conditional rate matrices

$$q_{\boldsymbol{a},\boldsymbol{b}} = \sum_{i=1}^{M} \left( q^{i|\,\mathrm{Par}(i)}_{a_i,b_i|\,\mathrm{P}_i(\boldsymbol{a})} \prod_{j \neq i} \delta_{a_j,b_j} \right). \tag{2}$$

where $\mathrm{P}_i(\boldsymbol{a})$ is a projection operator that project a complete assignment $\boldsymbol{a}$ to an assignment to the $\mathrm{Par}(i)$ components. Eq. (2) is, using the terminology of Nodelman et al. (2002), the "amalgamation" of the $M$ conditional rate matrices. Note the compact representation, which is valid for both diagonal and off-diagonal entries. It is also noteworthy that amalgamation is a summation, rather than a product; indeed, independent exponential rates are additive. If, for example, every component has $d$ possible values and $k$ parents, the rate matrix requires only $Md^{k+1}(d-1)$ parameters, rather than $d^M(d^M-1)$.

The dependency relations between components can be represented graphically as a directed graph, $\mathcal{G}$, in which each node corresponds to a component, and each directed edge defines a parent-child relation. A CTBN consists of such a graph, supplemented with a set of $M$ conditional rate matrices $\mathbb{Q}^{i|\,\mathrm{Par}(i)}$. The graph structure has two main roles: (i) it provides a data structure to which parameters are associated; (ii) it provides a qualitative description of dependencies among the various components of the system. The graph structure also reveals conditional independencies between sets of components (Nodelman et al., 2002).

Notational conventions: Full trajectories and observed pointwise values of components are denoted by lower case letters indexed by the relevant time intervals, e.g., $x_i^{(t)}$, $x_i^{[s,t]}$. We will use $\Pr(x_i^{(t)})$ and $\Pr(x_i^{[s,t]})$ as shorthands for $\Pr(X_i^{(t)} = x_i^{(t)})$ and $\Pr(X_i^{[s,t]} = x_i^{[s,t]})$.

It should be emphasized that even though CTBNs provide a succinct representation of multi-component processes, any inference query still requires the exponentiation of the full $d^M \times d^M$ dimensional rate matrix $\mathbb{Q}$. For example, given the state of the system at times $0$ and $T$, the *Markov bridge* formula is

$$\Pr(\boldsymbol{X}^{(t)} = \boldsymbol{a}|\boldsymbol{x}^{(0)}, \boldsymbol{x}^{(T)}) = \frac{[\exp(t\mathbb{Q})]_{\boldsymbol{x}^{(0)},\boldsymbol{a}}[\exp((T-t)\mathbb{Q})]_{\boldsymbol{a},\boldsymbol{x}^{(T)}}}{[\exp(T\mathbb{Q})]_{\boldsymbol{x}^{(0)},\boldsymbol{x}^{(T)}}}.$$

It is the premise of this work that such expressions cannot be computed directly, thus requiring approximation algorithms.

# 3 Sampling in a Two Component Process

## 3.1 Introduction

We will start by addressing the task of sampling from a two components process. The generalization to multi-component processes will follow in the next section.

Consider a two-component CTBN, $\boldsymbol{X} = (X, Y)$, whose dynamics is defined by conditional rates $\mathbb{Q}^{X|Y}$ and $\mathbb{Q}^{Y|X}$ (that is, $X$ is a parent of $Y$ and $Y$ is a parent of $X$). Suppose that we are given partial evidence about the state of the system. This evidence might contain point observations, as well as continuous observations in some intervals, of the states of one or two components. Our goal is to sample a trajectory of $(X, Y)$ from the joint posterior distribution.

The approach we take here is to use a Gibbs sampler (Gilks et al., 1996) over trajectories. In such a sampler, we initialize $X$ and $Y$ with trajectories that are consistent with the evidence. Then, we randomly either sample a trajectory of $X$ given the entire trajectory of $Y$ and the evidence on $X$, or sample a trajectory of $Y$ given the entire trajectory of $X$ and the evidence on $Y$. This procedure defines a random walk in the space of $(X, Y)$ trajectories. The basic theory of Gibbs sampling suggests that this random walk will converge to the distribution of $X, Y$ given the evidence.

To implement such a sampler, we need to be able to sample the trajectory of one component given the entire trajectory of the other component and the evidence. Suppose, we have a fully observed trajectory on $Y$. In this case, observations on $X$ at the extremities of some time interval statistically separate this interval from the rest of trajectory. Thus, we can restrict our analysis to the following situation: the process is restricted to a time interval $[0, T]$ and we are given observations $X^{(0)} = x^{(0)}$ and $X^{(T)} = x^{(T)}$, along with the entire trajectory of $Y$ in $[0, T]$. The latter consists of a sequence of states $(y_0, \ldots, y_K)$ and transition times $(\tau_0 = 0, \tau_1, \ldots, \tau_K, \tau_{K+1} = T)$. An example of such scenario is shown in Figure 1(a). The entire problem is now reduced to the following question: how can we sample a trajectory of $X$ in the interval $(0, T)$ from its posterior distribution?

To approach this problem we exploit the fact that *the sub-process $X$ given that $Y^{[0,T]} = y^{[0,T]}$ is Markovian* (although non-homogeneous in time):

**Proposition 3.1:** *The following Markov property holds for all $t > s$,*

$$\Pr(X^{(t)} \mid x^{[0,s]}, x^{(T)}, y^{[0,T]}) = \Pr(X^{(t)} \mid x^{(s)}, x^{(T)}, y^{[s,T]}).$$

## 3.2 Time Granularized Process

Analysis of such process requires reasoning about a continuum of random variables. A natural way of doing so is to perform the analysis in discrete time with a finite time granularity $h$, and examine the behavior of the system when we take $h \downarrow 0$.

To do so, we introduce some definitions. Suppose $\Pr$ is the probability function associated with a continuous-time Markov process with rate matrix $\mathbb{Q}$. We define the $h$-*coarsening* of $\Pr$ to be $\Pr_h$, a distribution over the random variables $\boldsymbol{X}^{(0)}, \boldsymbol{X}^{(h)}, \boldsymbol{X}^{(2h)}, \ldots$ which is defined by the dynamics

$$\Pr_h(\boldsymbol{X}^{(t+h)} = \boldsymbol{b} \mid \boldsymbol{X}^{(t)} = \boldsymbol{a}) = \delta_{\boldsymbol{a},\boldsymbol{b}} + h \cdot q_{\boldsymbol{a},\boldsymbol{b}},$$

which is the Taylor expansion of $[\exp(t\mathbb{Q})]_{\boldsymbol{a},\boldsymbol{b}}$, truncated at the linear term. When $h < \min_{\boldsymbol{a}}(-1/q_{\boldsymbol{a},\boldsymbol{a}})$, $\Pr_h$ is a well-defined distribution.

We would like to show that the measure $\Pr_h(A)$ of an event $A$ converges to $\Pr(A)$ when $h \downarrow 0$. To do so, however, we need to define the $h$-coarsening of an event. Given a time point $t$, define $\lfloor t \rfloor_h$ and $\lceil t \rceil_h$ to be the rounding down and up of $t$ to the nearest multiple of $h$. For point events we define $[\![\boldsymbol{X}^{(t)} = \boldsymbol{a}]\!]_h$ to be the event $\boldsymbol{X}^{(\lfloor t \rfloor_h)} = \boldsymbol{a}$, and $[\![\boldsymbol{X}^{(t^+)} = \boldsymbol{a}]\!]_h$ to the event $\boldsymbol{X}^{(\lceil t \rceil_h)} = \boldsymbol{a}$. For an interval event, we define $[\![\boldsymbol{X}^{(s,t]} = \boldsymbol{a}_{(s,t]}]\!]_h$ to be the event $\boldsymbol{X}^{(\lceil s \rceil_h)} = \boldsymbol{a}_{\lceil s \rceil_h}, \boldsymbol{X}^{(\lceil s \rceil_h + h)} = \boldsymbol{a}_{\lceil s \rceil_h + h}, \ldots, \boldsymbol{X}^{(\lfloor t \rfloor_h)} = \boldsymbol{a}_{\lfloor t \rfloor_h}$. Similarly, we can define the coarsening of events over only one component and composite events.

Note that the probability of any given trajectory tends to zero as $h \rightarrow 0$. The difficulty in working directly in the continuous-time formulation is that we condition on events that have zero probability. The introduction of a granularized process allows us to manipulate well-defined conditional probabilities, which remain finite as $h \rightarrow 0$.

**Theorem 3.2:** Let $A$ and $B$ be point, interval, or a finite combination of such events. Then

$$\lim_{h \downarrow 0} \Pr_h([\![A]\!]_h \mid [\![B]\!]_h) = \Pr(A \mid B)$$

From now on, we will drop the $[\![A]\!]_h$ notation, and assume it implicitly in the scope of $\Pr_h()$.

A simple minded approach to solve our problem is to work with a given finite $h$ and use discrete sampling to sample trajectories in the coarsened model (thus, working with a *dynamical Bayesian network*). If $h$ is sufficiently small this might be a reasonable approximation to the desired distribution. However, this approach suffers from sub-optimality due to this fixed time granularity — a too coarse granularity leads to inaccuracies, while a too fine granularity leads to computational overhead. Moreover, when different components evolve at different rates, this trade-off is governed by the fastest component.

### 3.3 Sampling a Continuous-Time Trajectory

To avoid the trade-offs of fixed time granularity we exploit the fact that while a single trajectory is defined over infinite time points, it involves only a finite number of transitions in a finite interval. Therefore, instead of sampling states at different time points, we only sample a finite sequence of transitions. The Markovian property of the conditional process $X$ enables doing so using a sequential procedure.

Our procedure starts by sampling the first transition time. It then samples the new state the transition leads to. As this new sample point statistically separates the remaining interval from the past, we are back with the initial problem yet with a shorter interval. We repeat these steps until the entire trajectory is sampled; it terminates once the next transition time is past the end of the interval.

Our task is to sample the first transition time and the next state, conditioned on $X^{(0)} = x^{(0)}$, $X^{(T)} = x^{(T)}$ as well as the entire trajectory of $Y$ in $[0,T]$. To sample this transition time, we first define the conditional cumulative distribution function $F(t)$ that $X$ stays in the initial state for a time less than $t$:

$$F(t) = 1 - \Pr\left(X^{(0,t]} = x^{(0)} | x^{(0)}, x^{(T)}, y^{[0,T]}\right) \qquad (3)$$

If we can evaluate this function, then we can sample the first transition time $\tau$ by inverse transform sampling — we draw $\xi$ from a uniform distribution in the interval $[0,1]$, and set $\tau = F^{-1}(\xi)$; see Figure 1a,b.

The Markov property of the conditional process allows us to decompose the probability that $X$ remains in its initial state until time $t$. Denoting the probability of $Y$'s trajectory and of $X$ remaining in its initial state until time $t$ by

$$p^{\text{past}}(t) = \Pr(X^{(0,t]} = x^{(0)}, y^{(0,t]} | x^{(0)}, y^{(0)}),$$

and the probability of future observations given the state of $(X_t, Y_t)$ by

$$p_x^{\text{future}}(t) = \Pr(x^{(T)}, y^{(t,T]} | X^{(t)} = x, y^{(t)}).$$

We can then write the probability that $X$ is in state $x^{(0)}$ until $t$ as

$$\Pr\left(X^{(0,t]} = x^{(0)} | x^{(0)}, x^{(T)}, y^{[0,T]}\right) = \frac{p^{\text{past}}(t) \cdot p_{x^{(0)}}^{\text{future}}(t)}{p_{x^{(0)}}^{\text{future}}(0)}. \qquad (4)$$

Lamentably, while the reasoning we just described is seemingly correct, all the terms in Eq. (4) are equal to 0, since they account for the probability of $Y$'s trajectory. However, as we shall see, if we evaluate this equation carefully we will be able to define it with terms that decompose the problem in a similar manner.

To efficiently compute these terms we exploit the fact that although the process is not homogeneous, the dynamics of the joint process within an interval $[\tau_k, \tau_{k+1})$ , in which $Y$ has a fixed value $y_k$, is characterized by a *single* unnormalized rate matrix whose entries depend on $y_k$. This allows us to adopt a *forward-backward* propagation scheme. We now develop the details of these propagations.

### 3.4 Computing $p^{\text{past}}(t)$

We begin with expressing $p^{\text{past}}(t)$ as a product of local terms. Recall that $p^{\text{past}}(t)$ is the probability that $X$ is constant until time $t$. We denote by $p_h^{\text{past}}(t)$ the $h$-coarsened version of $p^{\text{past}}(t)$.

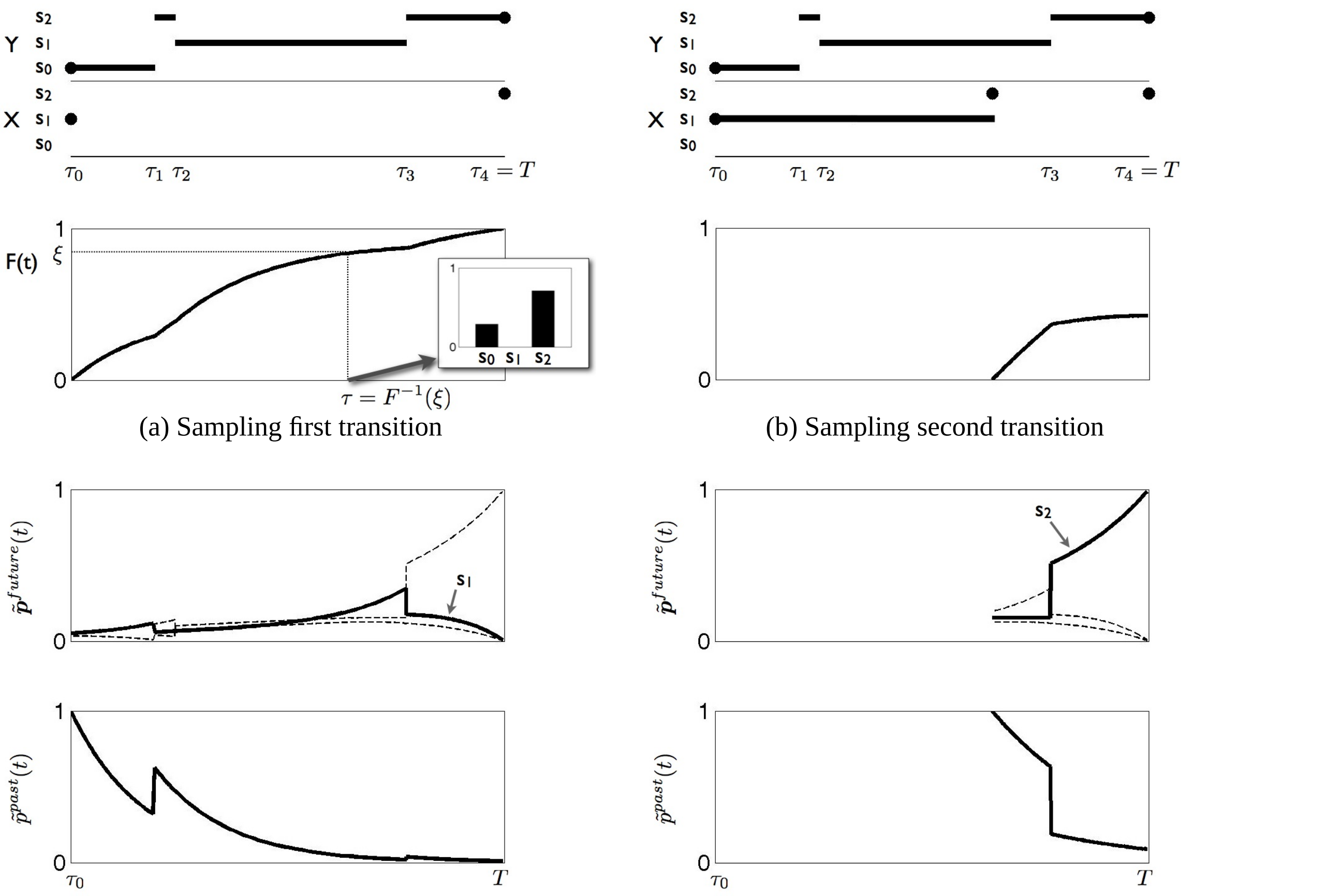

(a) Sampling first transition

(b) Sampling second transition

(c) Initial propagators

(d) Propagators in second step

Figure 1: Illustration of sampling of a single component with three states. (a) Top panel: sampling scenario, with a complete trajectory for $Y$, that has four transitions at $\tau_1, \ldots, \tau_4$, and point evidence on $X$ at times 0 and $T$. Bottom panel: the cumulative distribution $F(t)$, that $X$ changes states before time $t$ given this evidence. We sample the next transition time by drawing $\xi$ from a uniform distribution and setting $\tau = F^{-1}(\xi)$. Note that as $x^{(0)} \neq x^{(T)}$, $F(T) = 1$. The bar graph represents the conditional distribution of the next state, given a transition at time $\tau$. (b) Same sampling procedure for the second transition. Here $F(T) < 1$ since it is not necessary for $X$ to change its state. (c and d) The two components used in computing $1 - F(t)$: $\tilde{p}^{\text{past}}(t)$ the probability that $X$ stays with a constant value until time $t$ and $Y$ has the observed trajectory until this time; and $\tilde{p}_t^{\text{future}}(x)$ the probability that $X$ transition's from state $x$ at $t$ to its observed state at time $T$ and $Y$ follows its trajectory from $t$ to $T$.

To characterize the dynamics within intervals $(\tau_k, \tau_{k+1})$ we define *constant propagator functions*

$$\phi^y_{h,x}(\Delta t) = \Pr_h(X^{(t,t+\Delta t]} = x, Y^{(t,t+\Delta t]} = y | X^{(t)} = x, Y^{(t)} = y)$$

These functions determine the probability that $X = x$ and $Y = y$ throughout an interval of length $\Delta t$ if they start with these values.

At time $\tau_{k+1}$ the $Y$ component changes it value from $y_k$ to $y_{k+1}$. The transition probability at this point is $h \cdot q^{Y|X}_{y_k,y_{k+1}|x^{(0)}}$. Thus, from the Markov property of the joint process it follows that for $t \in (\tau_k, \tau_{k+1})$

$$p^{\text{past}}_h(t) = \left[\prod_{l=0}^{k-1} \phi^{y_l}_{h,x^{(0)}}(\Delta_l) \cdot q^{Y|X}_{y_l,y_{l+1}|x^{(0)}} \cdot h\right] \phi^{y_k}_{h,x^{(0)}}(t - \tau_k)$$

where $\Delta_l = \tau_{l+1} - \tau_l$.

To compute the constant propagator functions, we realize that in each step within the interval $(s, t]$ the state does not change. Thus,

$$\phi^y_{h,x}(\Delta t) = [1 + h \cdot (q^{X|Y}_{x,x|y} + q^{Y|X}_{y,y|x})]^{\frac{\lfloor \Delta t \rfloor_h}{h}}$$

We define

$$\phi^y_x(\Delta t) = \lim_{h \downarrow 0} \phi^y_{h,x}(\Delta t) = e^{(\Delta t)(q^{X|Y}_{x,x|y} + q^{Y|X}_{y,y|x})}$$

We conclude that if

$$\tilde{p}^{\text{past}}(t) = \left[\prod_{l=0}^{k-1} \phi^{y_l}_{x^{(0)}}(\Delta_l) \cdot q^{Y|X}_{y_l,y_{l+1}|x^{(0)}}\right] \phi^{y_k}_{x^{(0)}}(t-\tau_k),$$

then for $t \in (\tau_k, \tau_{k+1})$

$$\lim_{h \downarrow 0} \frac{p^{\text{past}}_h(t)}{h^k} = \tilde{p}^{\text{past}}(t)$$

### 3.5 Computing $p^{\textbf{future}}_x(t)$

We now turn to computing $p^{\text{future}}_x(t)$. Unlike the previous case, here we need to compute this term for every possible value of $x$. We do so by backward dynamic programing (reminiscent of backward messages in HMMs).

We denote by $\boldsymbol{p}^{\text{future}}_h(t)$ a vector with entries $p^{\text{future}}_{h,x}(t)$. Note that, $\boldsymbol{p}^{\text{future}}_h(T) = \boldsymbol{e}_{x^{(T)}}$ where $\boldsymbol{e}_x$ is the unit vector with 1 in position $x$. Next, we define a *propagator matrix* $\mathbb{S}^y_h(\Delta t)$ with entries

$$s^y_{h,a,b}(\Delta t) = \Pr_h(X^{(t+\Delta t)} = b, Y^{(t,t+\Delta t]} = y | X^{(t)} = a, Y^{(t)} = y)$$

This matrix provides the dynamics of $X$ in an interval where $Y$ is constant. We can use it to compute the probability of transitions between states of $X$ in the intervals $(\tau_k, \tau_{k+1}]$, for every $\tau_k < s < t < \tau_{k+1}$

$$\boldsymbol{p}^{\text{future}}_h(s) = \mathbb{S}^{y_k}_h(t-s)\boldsymbol{p}^{\text{future}}_h(t)$$

At transition points $\tau_k$ we need to take into account the probability of a change. To account for such transitions, we define a diagonal matrix $\mathbb{T}^{y,y'}$ whose $(a,a)$ entry is $q^{Y|X}_{y,y'|a}$. Using this notation and the Markov property of the joint process the conditional probability of future observations for $\tau_k \le t \le \tau_{k+1}$ is

$$\boldsymbol{p}^{\text{future}}_h(t) = \mathbb{S}^{y_{k-1}}_h(\tau_{k+1}-t)\left[\prod_{l=k+1}^{K} h\mathbb{T}^{y_l,y_{l+1}}\mathbb{S}^y_h(\Delta_l)\right]\boldsymbol{e}_{x^{(T)}}$$

It remains to determine the form of the propagator matrix. At time granularity $h$, we can write the probability of transitions between states of $X$ while $Y = y$ as a product of transition matrices. Thus,

$$\mathbb{S}^y_h(\Delta t) = (I + h \cdot \mathbb{R}_{X|y})^{\frac{\lfloor \Delta t \rfloor_h}{h}}$$

where $\mathbb{R}^{X|y}$ is the matrix with entries

$$r^{X|y}_{a,b} = \begin{cases} q^{X|Y}_{a,b|y} & a \ne b \\ q^{X|Y}_{a,a|y} + q^{Y|X}_{y,y|a} & a = b \end{cases}$$

We now can define

$$\mathbb{S}^y(\Delta t) = \lim_{h \downarrow 0} \mathbb{S}^y_h(\Delta t) = e^{(\Delta t)\mathbb{R}^{X|y}}$$

This terms is similar to transition matrix of a Markov process. Note, however that $\mathbb{R}$ is not a stochastic rate matrix, as the rows do not sum up to 0. In fact, the sum of the rows in negative, which implies that the entries in $\mathbb{S}^y_h(\Delta t)$ tend to get smaller with $\Delta t$. This matches the intuition that this term should capture the probability of the evidence that $Y = y$ for the whole interval.

To summarize, if we define for $t \in (\tau_k, \tau_{k+1})$

$$\tilde{\boldsymbol{p}}^{\text{future}}(t) = \mathbb{S}^{y_{k-1}}(\tau_{k+1}-t)\left[\prod_{l=k+1}^{K} \mathbb{T}^{y_l,y_{l+1}}\mathbb{S}^y(\Delta_l)\right]\boldsymbol{e}_{x^{(T)}},$$

then

$$\lim_{h \downarrow 0} \frac{\boldsymbol{p}^{\text{future}}_h(t)}{h^{K-k}} = \tilde{\boldsymbol{p}}^{\text{future}}(t)$$

### 3.6 Putting it All Together

Based on the above arguments.

$$\Pr_h\left(X^{(0,t]} = x^{(0)} | x^{(0)}, x^{(T)}, y^{[0,T]}\right) = \frac{p^{\text{past}}_h(t) p^{\text{future}}_{h,x^{(0)}}(t)}{p^{\text{future}}_{h,x^{(0)}}(0)}$$

Now, if $t \in (\tau_k, \tau_{k+1})$, then

$$\begin{aligned} &\Pr\left(X^{(0,t]} = x^{(0)} | x^{(0)}, x^{(T)}, y^{[0,T]}\right) \\ &= \lim_{h \downarrow 0} \frac{p^{\text{past}}_h(t) p^{\text{future}}_{h,x^{(0)}}(t)}{p^{\text{future}}_{h,x^{(0)}}(0)} \\ &= \lim_{h \downarrow 0} \frac{[h^{-k} p^{\text{past}}_h(t)][h^{-(K-k)} p^{\text{future}}_{h,x^{(0)}}(t)]}{h^{-K} p^{\text{future}}_{h,x^{(0)}}(0)} \\ &= \frac{\tilde{p}^{\text{past}}(t) \tilde{p}^{\text{future}}_{x^{(0)}}(t)}{\tilde{p}^{\text{future}}_{x^{(0)}}(0)} \end{aligned}$$

Thus, in both numerator and denominator we must account for the observation of $K$ transitions of $Y$, which have probability of $o(h^K)$. Since these term cancels out, we remain with the conditional probability over the event of interest.

### 3.7 Forward Sampling

To sample an entire trajectory we first compute $\tilde{\boldsymbol{p}}^{\text{future}}(t)$ only at transition times from the final transition to the start.

We sample the first transition time by drawing a random value $\xi$ from a uniform distribution in $[0,1]$. Now we find $\tau$ such that $F(\tau) = \xi$ in two steps: First, we sequentially search for the interval $[\tau_k, \tau_{k+1}]$ such that $F(\tau_k) \le F(\tau) \le F(\tau_{k+1})$ by propagating $\tilde{p}^{\text{past}}(t)$ forward through transition points. Second, we search the exact time point within $[\tau_k, \tau_{k+1}]$ using binary search with $L$ steps to obtain accuracy of $2^{-L}\Delta_k$. This step requires computation of $\mathbb{S}^{y_k}(2^{-L}\Delta_k)$ and its exponents $\mathbb{S}^{y_k}(2^{-l}\Delta_k)$, $l = 1, \ldots, L-1$.

Once we sample the transition time $t$, we need to compute the probability of the new state of $X$. Using similar arguments as the ones we discussed above, we find that

$$\Pr\left(X^{(t^+)} = x | X^{[0,t)} = x^{(0)}, X^{(t^+)} \neq x^{(0)}, y^{[0,T]}\right) = \frac{q^{X|Y}_{x^{(0)},x} \cdot \tilde{p}^{\text{future}}_x(t)}{\sum_{x' \neq x^{(0)}} q^{X|Y}_{x^{(0)},x'} \cdot \tilde{p}^{\text{future}}_{x'}(t)}.$$

Thus, we can sample the next state by using the pre-computed value of $\tilde{p}^{\text{future}}_x(t)$ at $t$.

Once we sample a transition (time and state), we can sample the next transition in the interval $[\tau, T]$. The procedure proceeds while exploiting propagators which have already been computed. It stops when $F(T) < \xi$, i.e., the next sampled transition time is greater than $T$. Figure 1 illustrates the conditional distributions of the first two transitions.

# 4 Sampling in a Multi-Component Process

The generalization from a two-component process to a general one is relatively straightforward. At each step, we need to sample a single component $X_i$ conditioned on trajectories in $\boldsymbol{Y} = (X_1, \ldots, X_{i-1}, X_{i+1}, \ldots, X_M)$. To save computations we exploit the fact that given complete trajectories over the Markov blanket of $X_i$, which is the component set of $X_i$'s parents, children and its children's parents, the dynamics in $X_i$ is independent of the dynamics of all other components (Nodelman et al., 2002).

Indeed, the structured representation of a CTBN allows computations using only terms involving the Markov blanket. To see that, we first notice that within an interval whose state is $\boldsymbol{Y}_t = \boldsymbol{y}$ the propagator matrix involves terms which depend only on the parents of $X_i$ $q^{X_i|Y}_{a,b|\boldsymbol{y}} = q^{X_i|\text{Par}(i)}_{a,b|\boldsymbol{u}_i}$ and terms which depend on the other members of the Markov blanket,

$$q^{\boldsymbol{Y}|X_i}_{\boldsymbol{y},\boldsymbol{y}|x_i} = \sum_{j \in \text{Child}(i)} q^{X_j|\text{Par}(j)}_{x_j,x_j|\boldsymbol{u}_j} + c_{\boldsymbol{y}}$$

where $c_{\boldsymbol{y}}$ does not depend on the state of $X_i$. Therefore, we define the reduced rate matrix $\mathbb{R}_{X_i|\boldsymbol{v}}$:

$$r^{X_i|\text{MB}(i)}_{a,b|\boldsymbol{v}} = \begin{cases} q^{X_i|\text{Par}(i)}_{a,b|\boldsymbol{u}_i} & a \neq b \\ q^{X_i|\text{Par}(i)}_{a,a|\boldsymbol{u}_i} + \sum_{j \in \text{Child}(i)} q^{X_j|\text{Par}(j)}_{x_j,x_j|\boldsymbol{u}_j} & a = b \end{cases}$$

where, $\boldsymbol{v}$ is the projection of $\boldsymbol{y}$ to the Markov blanket. Consequently the local propagator matrix becomes

$$\mathbb{S}^{\boldsymbol{v}}(t) = \exp(t \cdot \mathbb{R}_{X_i|\boldsymbol{v}}) \quad (5)$$

Importantly, this matrix differs from $\mathbb{S}^{\boldsymbol{y}}(t)$ by a scalar factor of $\exp(t \cdot c_{\boldsymbol{y}})$. The same factor arise when replacing the term in the exponent of the constant propagator. Therefore, these terms cancel out upon normalization.

This development also shows that when sampling $X_i$ we only care about transition points of one of the trajectories in $\text{MB}(i)$. Thus, the intervals computed in the initial backward propagation are defined by these transitions. Therefore, the complexity of the backward procedure scales with the rate of $X_i$ and its Markov blanket.

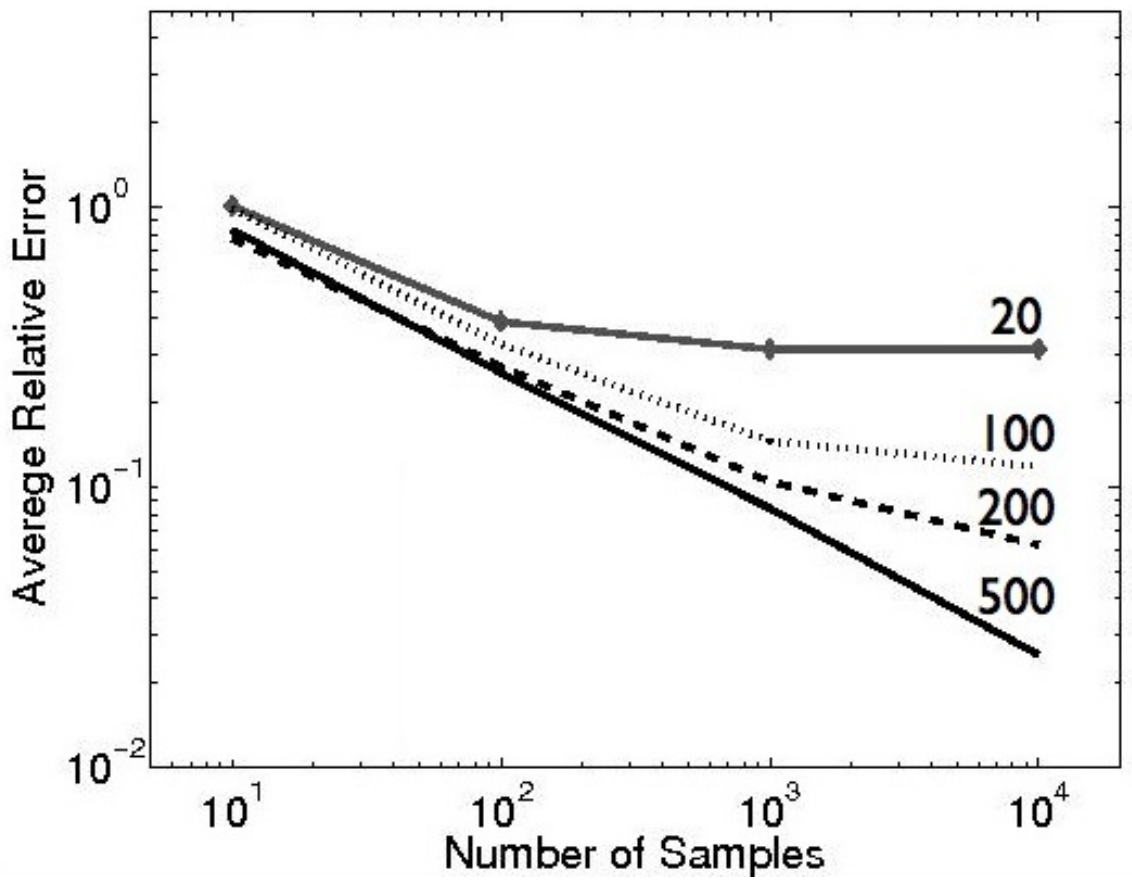

Figure 2: Relative error versus burn-in and number of samples.

# 5 Experimental Evaluation

We evaluate convergence properties of our procedure on a chain network presented in Fan and Shelton (2008), as well as on related networks of various sizes and parametrizations. The basic network contains 5 components, $X_0, \rightarrow X_1 \rightarrow \ldots X_4$, with 5 states each. The transition rates of $X_0$ suggest a tendency to cycle in 2 possible loops: $s_0 \rightarrow s_1 \rightarrow s_2 \rightarrow s_0$ and $s_0 \rightarrow s_3 \rightarrow s_4 \rightarrow s_0$; whereas for $i > 0$, $X_i$ attempts to follow the state of $X_{i-1}$ — the transition $q^{X_i|X_{i-1}}_{a,b|c}$ has higher intensity when $c = b$. The intensities of $X_0$ in the original network are symmetric relative to the two loops. We slightly perturbed parameters to break symmetry since the symmetry between the two loops tends to yield untypically fast convergence.

To obtain a reliable convergence assessment, we should generate samples from multiple independent chains which are initialized from an over-dispersed distribution. Aiming to construct such samples, our initialization procedure draws for each component a rate matrix by choosing an assignment to its parents from a uniform distribution and taking the corresponding conditional rate matrix. Using these matrices it samples a trajectory that is consistent with evidence independently for every component using the backward propagation-forward sampling strategy we described above.

A crucial issue in MCMC sampling is the time it takes the chain to *mix* — that is, sample from a distribution that is close to the target distribution rather than the initial distribution. It is not easy to show empirically that a chain has mixed. We examine this issue from a pragmatic perspective by asking what is the quality of the estimates based on sam-ples taken at different number of "burn-in" iterations after the initialization, where a single iteration involves sampling each of the components once. We examine the estimates of expected sufficient statistics that are required for learning CTBN's — residence time of components in states and the number of transitions given the state of the component's parent (Nodelman et al., 2003). We measure estimation quality by the *average relative error* $\sum_j \frac{|\hat{\theta}_j - \theta_j|}{\theta_j}$ where $\theta_j$ is exact value of the $j$'th sufficient statistics calculated using numerical integration and $\hat{\theta}_j$ is the approximation.

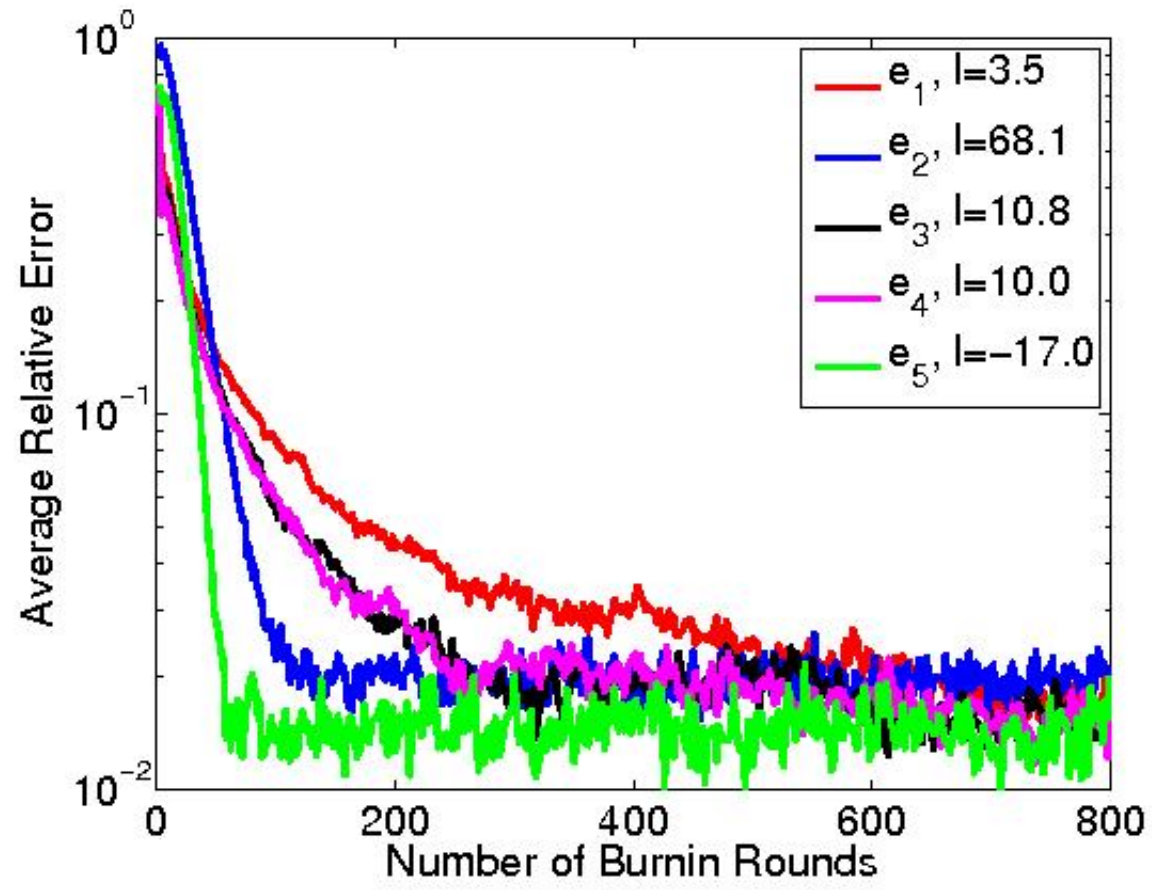

Figure 3: Error versus burn-in for different evidence sets. For each set we specify the average log-likelihood of the samples after convergence.

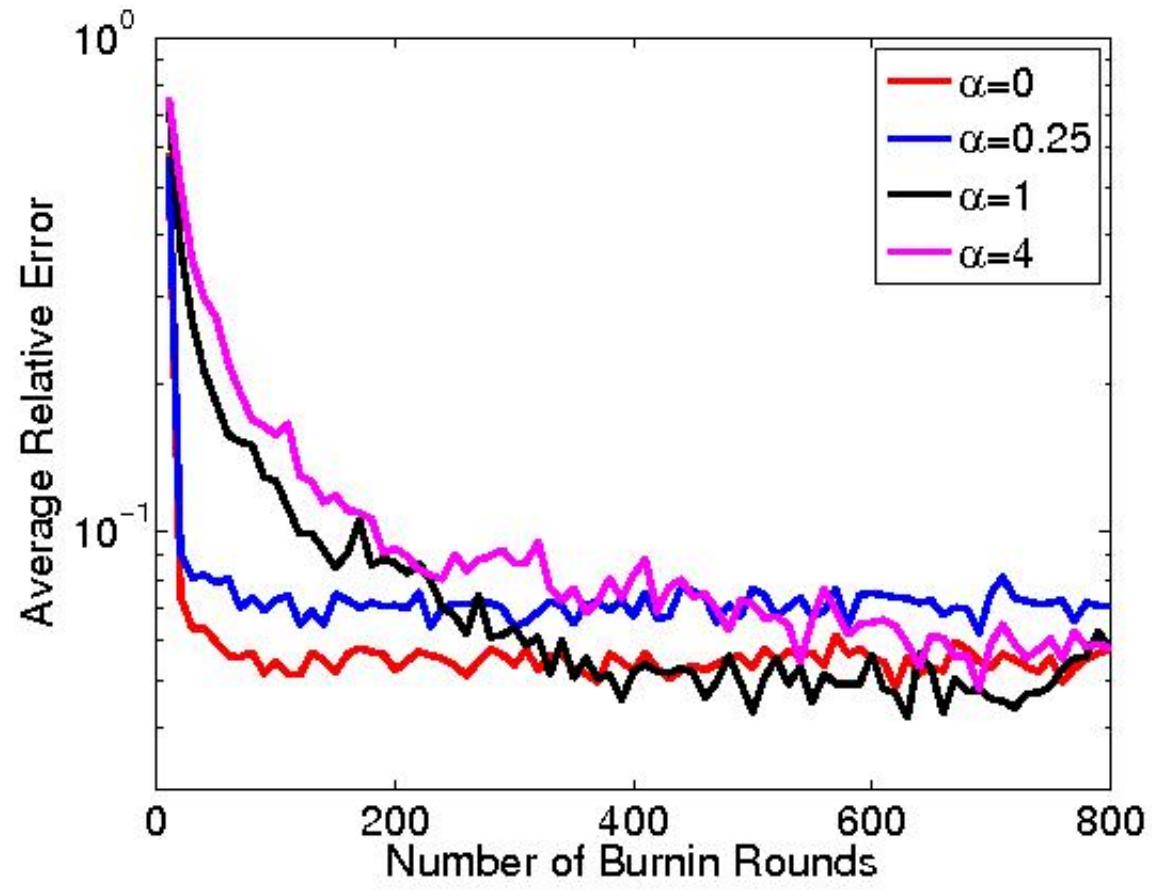

Figure 4: Effect of conditional transition probability sharpness on mixing time.

To make the task harder, we chose an extreme case by setting evidence $\boldsymbol{X}^{(0)} = \vec{s}_0$ (the vector of $s_0$), and $\boldsymbol{X}^{(3)} = (s_0, s_1, s_3, s_0, s_1)$. We then sampled the process using multiple random starting points, computed estimated expected statistics, and compared them the exact expected statistics. Figure 2 shows the behavior of the average relative error taken over all $\theta > 0.05$ versus the sample size for different number of burn-in iterations. Note that when using longer burn-in, the error decreases at a rate of $O(\sqrt{n})$, where $n$ is the number of samples, which is what we would expect from theory, if the samples where totally independent. This implies that at this long burn-in the error due to the sampling process is smaller than the error contributed by the number of samples.

To study further the effect of evidence's likelihood, we measured error versus burn-in using 10,000 samples in our original evidence set, and four additional ones. The first additional evidence, denoted by $\boldsymbol{e}_2$ is generated by setting $\boldsymbol{X}^{(0)} = \vec{s}_0$, forward sampling a random trajectory and taking the complete trajectory of $X_4$ as evidence. Additional sets are: $\boldsymbol{e}_3 = \{\boldsymbol{X}^{(0)} = \vec{s}_0, \boldsymbol{X}^{(3)} = \vec{s}_0\}$; $\boldsymbol{e}_4 = \{\boldsymbol{X}^{(0)} = \vec{s}_0\}$ and an extremely unlikely case $\boldsymbol{e}_5 = \{\boldsymbol{X}^{(0)} = \vec{s}_0, X_0^{(0,3)} = s_0, \boldsymbol{X}^{(3)} = (s_0, s_1, s_3, s_0, s_1)\}$.

Figure 3 illustrates that burn-in period may vary by an order of magnitude, however it is not correlated with the log-likelihood. Note that in this specific experiment slower convergence occurs when continuous evidence is absent. The reason for this may be the existence of multiple possible paths that cycle through state zero. That is, the posterior distribution is , in a sense, multi-modal.

To further explore the effect of the posterior's landscape, we tested networks with similar total rate of transitions, but with varying level of coupling between components. Stronger coupling of components leads to a sharper joint distribution. To achieve variations in the coupling we consider variants of the chain CTBN where we set $\hat{\pi}_{a,b|y} = \frac{(q_{a,b|y})^\alpha}{\sum_{c \neq a}(q_{a,c|y})^\alpha}$ and $\hat{q}_{a,b|y} = q_{a,a|y} \cdot \hat{\pi}_{a,b|y}$ where $\alpha$ is a non-negative sharpness parameter As $\alpha \rightarrow 0$ the network becomes smoother, which reduces coupling between components. However, the stationary distribution is not tending to a uniform one because we do not alter the diagonal elements. Figure 4 shows convergence behavior for different values of $\alpha$ where estimated statistics are averaged over 1,000 samplers. As we might expect, convergence is faster as the network becomes smoother.

Next we evaluated the scalability of the algorithm by generating networks containing additional components with an architecture similar to the basic chain network. As exact inference is infeasible in such networks we measured relative error versus estimations taken from long runs. Specifically, for each $N$, we generated 1000 samples by running 100 independent chains and taking samples after 10,000 rounds as well as additional 9 samples from each chain every 1,000 rounds. Using these samples we estimated the target sufficient statistics. To avoid averaging different numbers of components, we compared the relative error in the estimate of 5 components for networks of different sizes. Figure 5 shows the results of this experiment. As we can see, convergence rates decay moderately

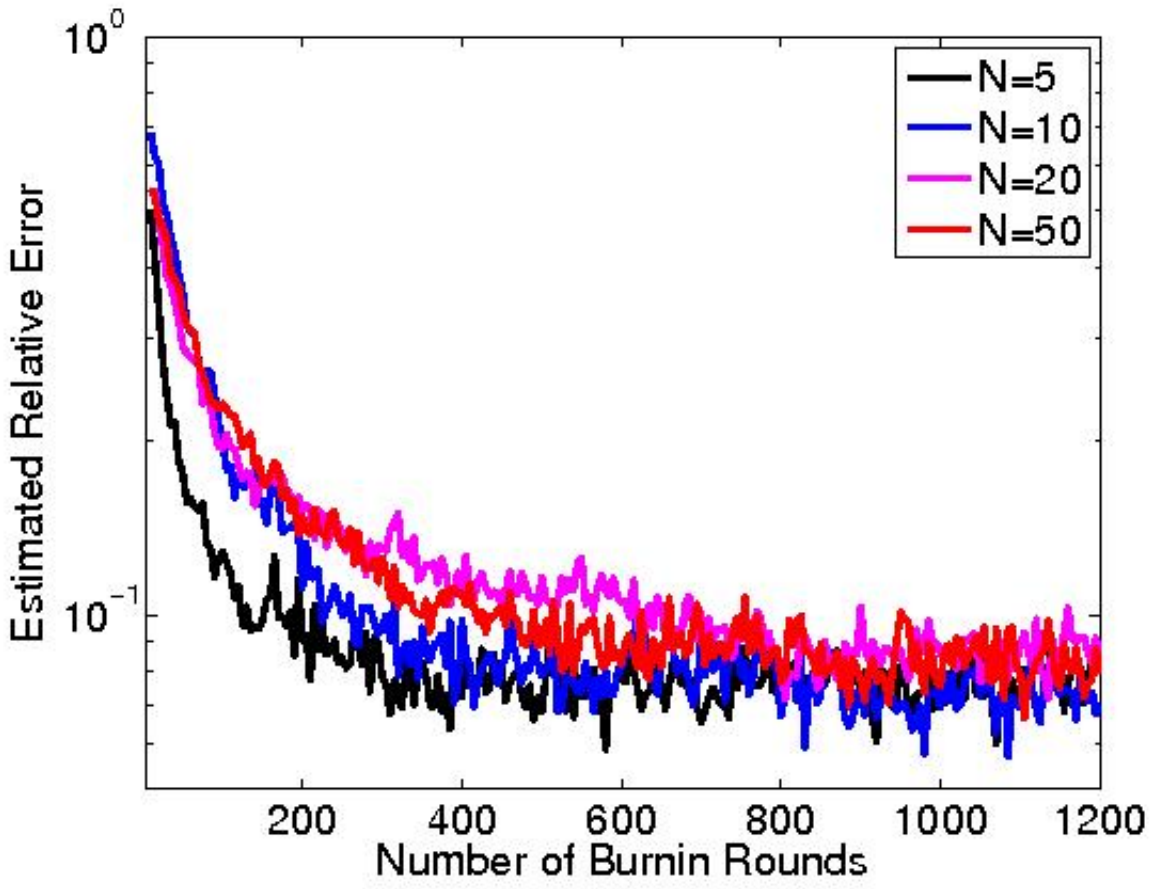

Figure 5: Convergence of relative error in statistics of first five components in networks of various sizes. Errors are computed with respect to statistics that are generated with $N = 10,000$ rounds.

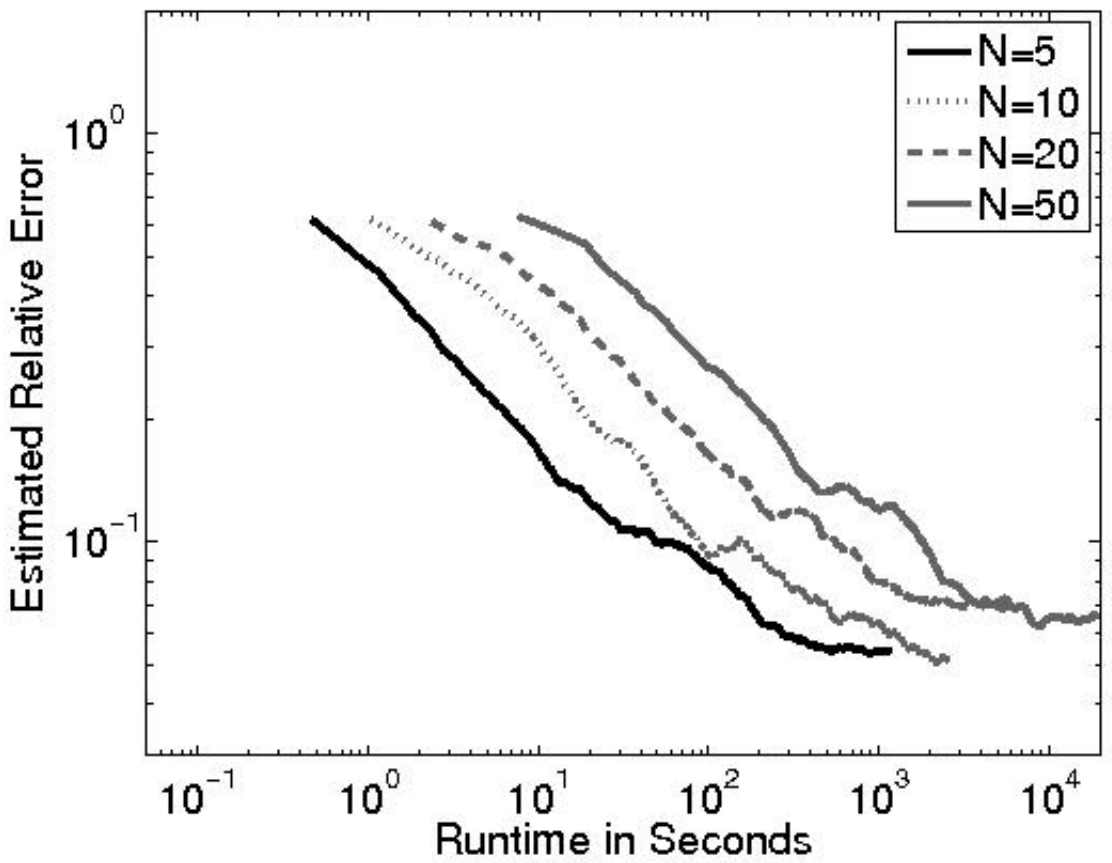

Figure 6: Relative error versus run-time in seconds for various network sizes.

with the size of the network.

While for experimental purposes we generate many samples independently. A practical strategy is to run a small number of chains in parallel and then collect take a large number of samples from each. We tested this strategy by generating 10 independent chain for various networks and estimating statistics from all samples except the first 20%. Using these, we measured how the behavior of error versus CPU run-time scales with network size. Average results of 9 independent tests are shown in Figure 6. Roughly, the run-time required for a certain level of accuracy scales linearly with network size.

Our sampling procedure is such that the cost of sampling a component depends on the time scales of its Markov neighbors and its own rate matrix. To demonstrate that, we created a chain network where each component has rates that are of half the magnitude of its parent. This means that the first component tends to switch state twice as fast as the second, the second is twice as fast as the third, and so on. When we examine the number of transitions in the sampled trajectories Figure 7, we see that indeed they are consistent with these rates, and quickly converge to the expected number, since in this example the evidence is relatively weak. When we examine the number of intervals in the Markov blanket of each components, again we see that neighbors of fast components have more intervals. In this graph $X_1$ is an anomaly since it does not have a parent.

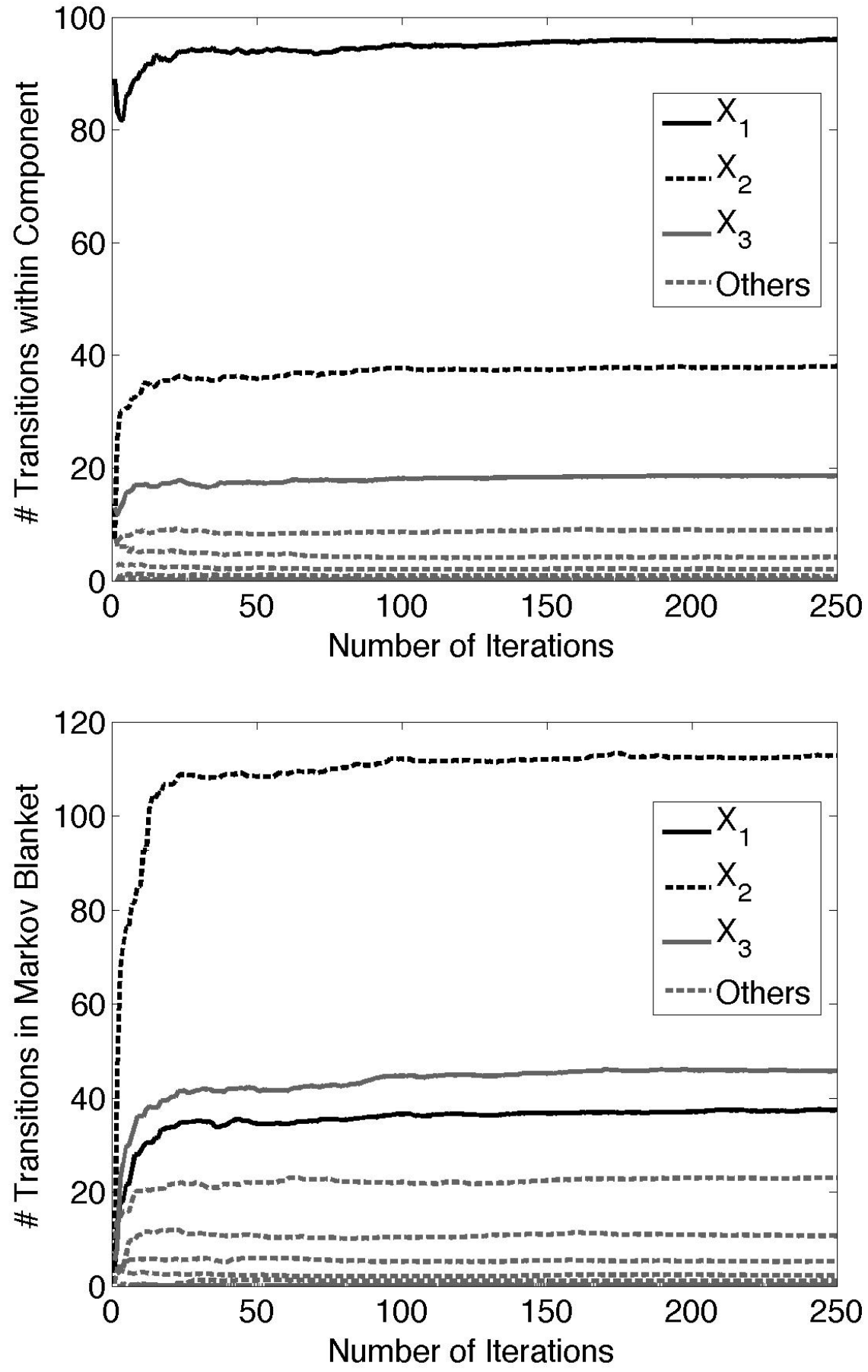

Figure 7: The effect of different time scales on the sampling. In this network $X_i$'s rate is twice as fast than $X_{i+1}$'s rate. (top) The number transitions sampled for each of the first four components as a function of iteration number. (bottom) The number of intervals of Markov neighbors of each component as a function of iteration number.

## 6 Discussion

In this paper we presented a new approach for approximate inference in Continuous-Time Bayesian Networks. By building on the strategy of Gibbs sampling. The core of our method is a new procedure for exact sampling of a trajectory of a single component, given evidence on its end points and the full trajectories of its Markov blanket components. This sampling procedure adapts in a natural way to the time scale of the component, and is exact, up to a predefined resolution, without sacrificing efficiency.

This is the first MCMC sampling procedure for this type of models. As such it provides an approach that can sample from the exact posterior, even for unlikely evidence. As the current portfolio of inference procedures for continuous-time processes is very small, our procedure provides another important tool for addressing these models. In particular, since the approach is *asymptotically unbiased* in the number of iterations it can be used to judge the systematic bias introduced by other, potentially faster, approximate inference methodologies, such as the one of Saria et al. (2007).

It is clear that sampling complete trajectories is not useful in situations where we expect a very large number of transitions in the relevant time periods. However, in many applications of interest, and in particular our long term goal of modeling sequence evolution (El-Hay et al., 2006), this is not the case. When one or few components transitions much faster than neighboring components, then we are essentially interested in its average behavior (Friedman and Kupferman, 2006). In such situations, it would be useful to develop a Rao-Blackwellized sampler that integrates over the fast components.

As with many MCMC procedures, one of the main concerns is the mixing time of the sampler. An important direction for future research is the examination of methods for accelerating the mixing - such as *Metropolis-coupled MCMC* or *simulated tempering* (Gilks et al., 1996) - as well as a better theoretic understanding of the convergence properties.

**Acknowledgments**

We thank Ido Cohn and the anonymous reviewers for helpful remarks on previous versions of the manuscript. This research was supported in part by grants from the Israel Science Foundation and the US-Israel Binational Science Foundation. Tal El-Hay is supported by the Eshkol fellowship from the Israeli Ministry of Science.

## References

Chung, K. (1960). *Markov chains with stationary transition probabilities*. Springer Verlag, Berlin.

El-Hay, T., Friedman, N., Koller, D., and Kupferman, R. (2006). Continuous time markov networks. In *Proceedings of the Twenty-second Conference on Uncertainty in AI (UAI)*.

Fan, Y. and Shelton, C. (2008). Sampling for approximate inference in continuous time Bayesian networks. In *Tenth International Symposium on Artificial Intelligence and Mathematics*.

Friedman, N. and Kupferman, R. (2006). Dimension reduction in singularly perturbed continuous-time Bayesian networks. In *Proceedings of the Twenty-second Conference on Uncertainty in AI (UAI)*.

Gardiner, C. (2004). *Handbook of stochastic methods*. Springer-Verlag, New-York, third edition.

Gikhman, I. and Skorokhod, A. (1975). *The theory of Stochastic processes II*. Springer Verlag, Berlin.

Gilks, W. R., S., R., and J., S. D. (1996). *Markov Chain Monte Carlo in Practice*. Chapman & Hall.

Heskes, T. and Zoeter, O. (2002). Expectation propagation for approximate inference in dynamic Bayesian networks. In *Uncertainty in Artificial Intelligence: Proceedings of the Eighteenth Conference (UAI-2002)*, pages 216–233.

Minka, T. P. (2001). Expectation propagation for approximate Bayesian inference. In *Proc. Seventeenth Conference on Uncertainty in Artificial Intelligence (UAI '01)*, pages 362–369.

Ng, B., Pfeffer, A., and Dearden, R. (2005). Continuous time particle filtering. In *Proceedings of the 19th International Joint Conference on AI*.

Nodelman, U., Shelton, C., and Koller, D. (2002). Continuous time Bayesian networks. In *Proc. Eighteenth Conference on Uncertainty in Artificial Intelligence (UAI '02)*, pages 378–387.

Nodelman, U., Shelton, C., and Koller, D. (2003). Learning continuous time Bayesian networks. In *Proc. Nineteenth Conference on Uncertainty in Artificial Intelligence (UAI '03)*, pages 451–458.

Nodelman, U., Shelton, C., and Koller, D. (2005). Expectation propagation for continuous time Bayesian networks. In *Proc. Twenty-first Conference on Uncertainty in Artificial Intelligence (UAI '05)*, pages 431–440.

Saria, S., Nodelman, U., and Koller, D. (2007). Reasoning at the right time granularity. In *Proceedings of the Twenty-third Conference on Uncertainty in AI (UAI)*.